\documentclass{article} % For LaTeX2e
\usepackage{iclr2015,times}
\usepackage{hyperref}
\usepackage{url}
\usepackage{subcaption}
\usepackage{graphicx}
\usepackage{amsmath}
\usepackage{amsfonts}
\usepackage{amssymb}
\graphicspath{ {./figs/} }

\title{Fully Convolutional Multi-Class Multiple Instance Learning}
\author{
Deepak Pathak, Evan Shelhamer, Jonathan Long \& Trevor Darrell\\
UC Berkeley\\
\texttt{\{pathak,shelhamer,jonlong,trevor\}@cs.berkeley.edu}
}

\iclrfinalcopy % Uncomment for camera-ready version

\begin{document}

\maketitle

\begin{abstract}
Multiple instance learning (MIL) can reduce the need for costly annotation in tasks such as semantic segmentation by weakening the required degree of supervision.
%Multiple instance learning (MIL) can help reduce the cost of strong annotations for supervised tasks such as semantic segmentation by weakening the supervision required.
%Multiple instance learning (MIL) can be of significant importance in tasks which require heavy annotations in supervised scenarios.
We propose a novel MIL formulation of multi-class semantic segmentation learning by a fully convolutional network.
%We propose a novel formulation of MIL in multi-class setting using fully convolutional networks.
In this setting, we seek to learn a semantic segmentation model from just weak image-level labels.
%We investigate the difficult problem of semantic image segmentation with just weak image-level labels.
The model is trained end-to-end to jointly optimize the representation while disambiguating the pixel-image label assignment.
Fully convolutional training accepts inputs of any size, does not need object proposal pre-processing, and offers a pixelwise loss map for selecting latent instances.
Our multi-class MIL loss exploits the further supervision given by images with multiple labels.
We evaluate this approach through preliminary experiments on the PASCAL VOC segmentation challenge.
\end{abstract}

\section{Introduction}

Convolutional networks (convnets) are achieving state-of-the-art performance on many computer vision tasks but require costly supervision.
Following the ILSVRC12-winning image classifier of \cite{alexnet}, progress on detection~\citep{rcnn} and segmentation~\citep{arxiv14_jonEvan} demonstrates that convnets can likewise address local tasks with structured output.

Most deep learning methods for these tasks rely on strongly annotated data that is highly time-consuming to collect.
Learning from weak supervision, though hard, would sidestep the annotation cost to scale up learning to available image-level labels.

In this work, we propose a novel framework for multiple instance learning (MIL) with a fully convolutional network (FCN).
The task is to learn pixel-level semantic segmentation from weak image-level labels that only signal the presence or absence of an object.
Images that are not centered on the labeled object or contain multiple objects make the problem more difficult.
% TODO has this really not been done? might want to phrase otherwise.
The insight of this work is to drive the joint learning of the convnet representation and pixel classifier by multiple instance learning.
%The key insight of this work is to exploit the deep network to jointly learn the representation as well as the pixel-level-classification.
Fully convolutional training learns the model end-to-end at each pixel.
To learn the segmentation model from image labels, we cast each image as a bag of pixel-level-instances and define a pixelwise, multi-class adaptation of MIL for the loss.

MIL can reduce the need for bounding box annotations~\citep{verbeek2014_cvpr,song2014weakly}, but it is rarely attempted for segmentation.
%MIL task of object detection to prevent the bounding box annotations~\citep{rcnn}, but seldom for segmentation.
\citet{oquab2014weakly} improve image classification by inferring latent object location, but do not evaluate the localization. \citet{lsdaMIL} train by MIL fine-tuning but rely on bounding box supervision and proposals for representation learning. Most MIL problems are framed as max-margin learning~\citep{andrews2002_nips,felzenszwalb2010_pami}, while other approaches use boosting~\citep{ali_cvpr14} or Noisy-OR models~\citep{heckerman2013tractable}.
%It has been framed mostly in max-margin paradigm~\citep{rcnn}, and sometimes in boosting framework~\citep{rcnn} or Noisy-OR settings~\citep{rcnn}.
These approaches are limited by (1) fixed representations and (2) sensitivity to initial hypotheses of the latent instance-level labels.
%These approaches suffer from two major drawbacks.
%First, they assume explicitly learned representations and dis-jointly optimize for the latent instance level information.
%Secondly, the alternating optimization procedure is extremely sensitive to the initial hypothesis for instance-level labels.
We aim to counter both shortcomings by simultaneously learning the representation to maximize the most confident inferred instances.
We incorporate multi-class annotations by making multi-class inferences for each image.
When an image / bag contains multiple classes the competition of pixelwise models help to better infer the latent instance-level classes.

% TODO bullet point summaries are redundant with the intro, so this could be dropped.
% TODO or could switch to the abbreviated list commented out below
We investigate the following ideas and carry out preliminary experiments to these ends:
\begin{itemize}
\item We perform MIL jointly with end-to-end representation learning in a fully convolutional network.
  This eliminates the need to instantiate instance-label hypotheses.
  FCN learning and inference can process images of different sizes without warping or object proposal pre-processing.
  This makes training simple and fast.
\item We propose a multi-class pixel-level loss inspired by the binary MIL scenario.
  This tries to maximize the classification score based on each pixel-instance, while simultaneously taking advantage of inter-class competition in narrowing down the instance hypotheses.
\item We target the under-studied problem of weakly supervised image segmentation.
  Our belief is that pixel-level consistency cues are helpful in disambiguating object presence.
  In this way weak segmentation can incorporate more image structure than bounding boxes.
\end{itemize}

% TODO could go back to the list
%%In this paper, we investigate the following things with the preliminary experiments :
%\begin{itemize}
%\item joint multiple instance and representation learning in a fully convolutional network trained end-to-end
%  %This eliminates the need of instantiation of instance-label hypothesis.
%  %FCN allows to process varying size images without warping, and thus eliminates the need for any object proposal based pre-processing during training, resulting into a fast and convenient training.
%\item a multi-class pixelwise loss inspired by the binary MIL setting; and
%  %This tries to maximize the classification score based on each pixel-instance, while simultaneously take advantage of their competence in narrowing down the instance hypothesis.
%\item evaluation of this model for weakly supervised semantic segmentation
%  %Our belief is that the pixel-level consistency cues are quite helpful in disambiguating the presence of object in image, thus weak segmentation utilizes more information of image structure than bounding boxes.
%\end{itemize}
%and carry out preliminary experiments to these ends.

\section{Fully Convolutional MIL}

A fully convolutional network (FCN) is a model designed for spatial prediction problems.
Every layer in an FCN computes a local operation on relative spatial coordinates.
In this way, an FCN can take an input of any size and produce an output of corresponding dimensions.

For weakly supervised MIL learning, the FCN allows for the efficient selection of training instances.
The FCN predicts an output map for all pixels, and has a corresponding loss map for all pixels.
This loss map can be masked, re-weighted, or otherwise manipulated to choose and select instances for computing the loss and back-propagation for learning.

We use the VGG 16-layer net~\citep{simonyan2014very} and cast it into fully convolutional form as suggested in~\citet{arxiv14_jonEvan} for semantic segmentation by replacing fully connected layers with corresponding convolutions.
The network is fine-tuned from the pre-trained ILSVRC classifier weights i.e. pre-trained to predict image-level labels.
We then experiment with and without initializing the last layer weights i.e. the classifier layer. These initializations, without MIL fine-tuning, act as the baselines (row 1 and 2 in Table). If there is no image-level pretraining, the model quickly converges to all background.
Semantic segmentation requires a background class but the classification task has none; we simply zero initialize the background classifier weights.

\section{Multi-Class MIL Loss}

%sigmoid entropy and softmax loss are the same right ?
%No, softmax is multinomial logistic loss = 1 of k while sigmoid cross entropy = l independent bernoullis.
%Ok, then wrire that it's multinomial logistic loss over the selected predictions and backpropagated through all layers.
%I'd re-arrange that 1st paragraph to start with the loss selected at maximum predictions, then state that the output map / loss map of the FCN lets you do this selection.

We define a multi-class MIL loss as the multi-class logistic loss computed at maximum predictions.
This selection is enabled by the output map produced by FCN i.e. for an image of any size, the FCN outputs a heat-map for each class (including background) of corresponding size.
We identify the max scoring pixel in the coarse heat-maps of classes present in image and background.
The loss is then only computed on these coarse points, and is back propagated through the network.
The alternating optimization in the binary MIL problem inspires this ignoring of the loss at non-maximally scoring points.
The background class is analogous to the negative instances by competing against the positive object classes.
Let the input image be $I$, its label set be $\mathcal{L}_I$ (including background label) and $\hat{p}_l(x,y)$ be the output heat-map for the $l^{th}$ label at location $(x,y)$. The loss is defined as:
\begin{align*}
(x_l,y_l) &= \arg \max_{\forall (x,y)} \hat{p}_l(x,y) \qquad \forall l\in\mathcal{L}_I\\
\implies \text{MIL LOSS }&=\frac{-1}{|\mathcal{L}_I|} \sum_{l\in\mathcal{L}_I} \log \hat{p}_l(x_l,y_l)
\end{align*}

Ignoring the loss at all non-maximally scoring points is key to avoid biasing the learning of the FCN to the background.
Simultaneous training exploits multi-label images through inter-class confusion to help refine the intra-class pixel accuracy.
At inference time, the MIL-FCN takes the top class prediction at every point in the coarse prediction and bilinearly interpolates to image resolution to obtain a pixelwise segmentation.

\section{Experiments}

All results are on the PASCAL VOC segmentation challenge.
We train and validate on the VOC 2011 train augmented by \cite{hariharan2011semantic} and val sets then evaluate on the completely held-out VOC 2012 test set.
The evaluation metric is intersection over union (IU), and is defined per class as the percentage of pixels in the intersection of ground truth segmentation mask, and the predicted mask out of the number of pixels in their union.
The MIL-FCN model is initialized from the 16-layer VGG ILSVRC14 classifier \citep{simonyan2014very} then fine-tuned by the MIL loss.
\cite{arxiv14_jonEvan} fine-tune from all but the output layer, as they have access to complete supervision.
In our setting however, transferring the output layer parameters for the classes common to both PASCAL and ILSVRC improves results.
Including these classifier parameters helps prevent degenerate solutions of predicting all background. We train our model with a learning rate $0.0001$ , momentum $0.9$ and weight decay $0.0005$.
The training is quick and the network converges in less than 10,000 iterations.

\begin{table}[h]
\caption{\emph{Results on PASCAL VOC 2011 segmentation validation and 2012 test data}. Fine-tuning with the MIL loss achieves 96\% relative improvement over the baseline.}
\label{table:result}
\begin{center}
\begin{tabular}{lcc}
\multicolumn{1}{c}{\bf Approach}  &\multicolumn{1}{c}{\bf mean IU (VOC2011 val)} &\multicolumn{1}{c}{\bf mean IU (VOC2012 test)}
\\ \hline \\
Baseline (no classifier) & 3.52\% & -\\
Baseline (with classifier) & 13.11\% & 13.09\%\\
\textbf{MIL-FCN} & \textbf{25.05}\% & \textbf{25.66}\%\\
Oracle (supervised) & 59.43\% & 63.80\%\\
\end{tabular}
\end{center}
\end{table}

\begin{figure}[h]
\begin{center}

\begin{subfigure}[b]{.49\linewidth}
    \centering
    \includegraphics[width=\textwidth,height=2.5cm]{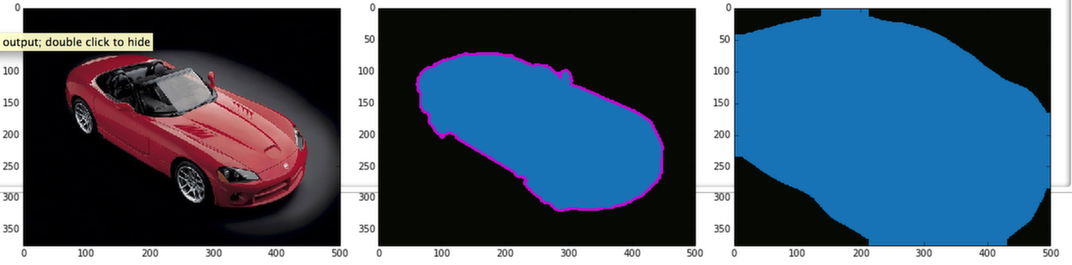}
\end{subfigure}\hfill
\begin{subfigure}[b]{.49\linewidth}
    \centering
    \includegraphics[width=\textwidth,height=2.5cm]{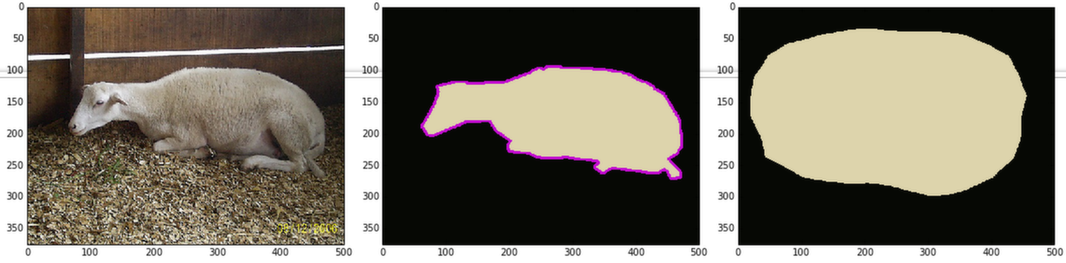}
\end{subfigure}\hfill

\begin{subfigure}[b]{.49\textwidth}
    \centering
    \includegraphics[width=\textwidth,height=2.5cm]{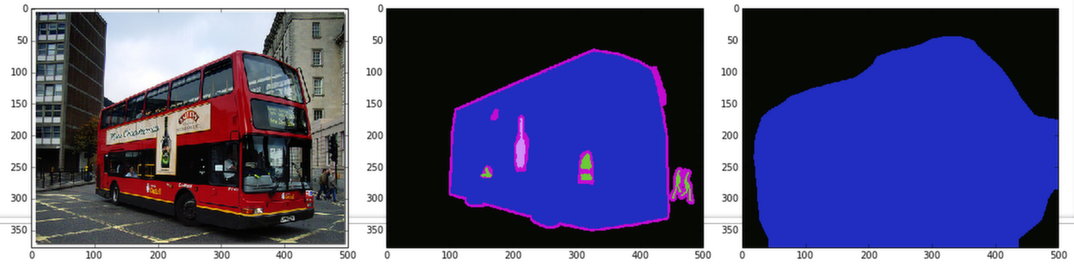}
\end{subfigure}\hfill
\begin{subfigure}[b]{.49\textwidth}
    \centering
    \includegraphics[width=\textwidth,height=2.5cm]{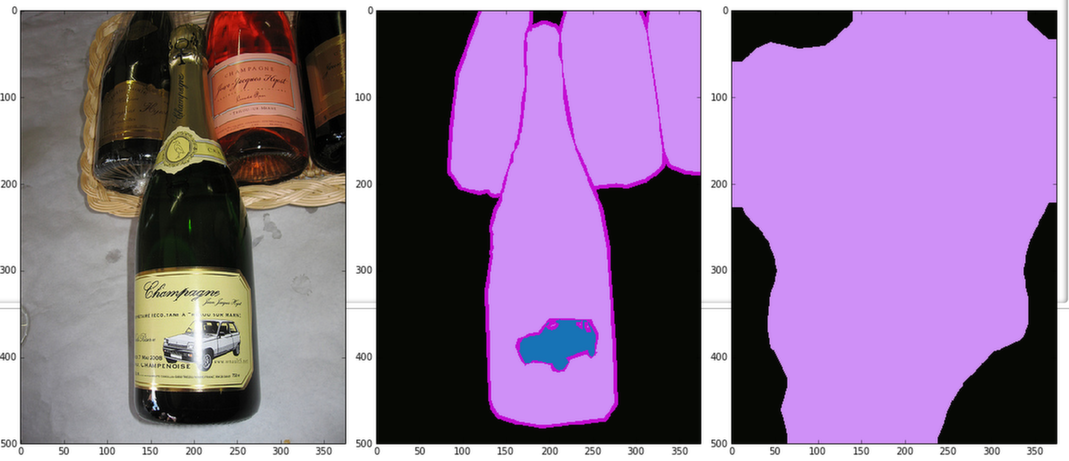}
\end{subfigure}\hfill

\begin{subfigure}[b]{.49\textwidth}
    \centering
    \includegraphics[width=\textwidth,height=2.5cm]{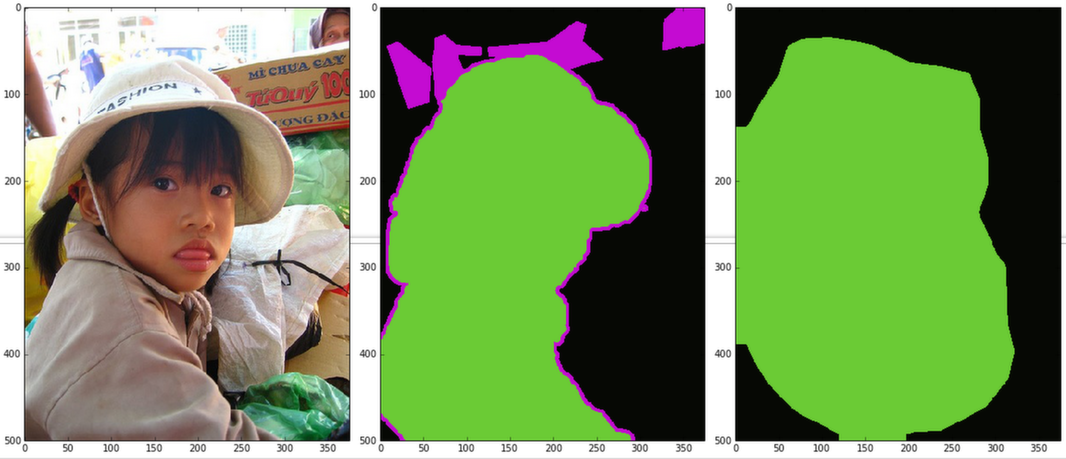}
\end{subfigure}\hfill
\begin{subfigure}[b]{.49\textwidth}
    \centering
    \includegraphics[width=\textwidth,height=2.5cm]{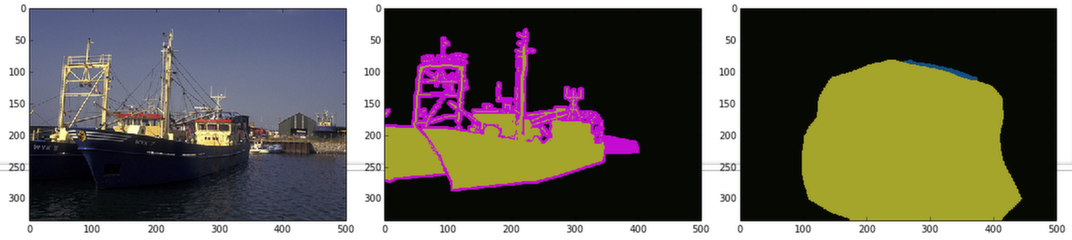}
\end{subfigure}\hfill

\begin{subfigure}[b]{.49\textwidth}
    \centering
    \includegraphics[width=\textwidth,height=2.5cm]{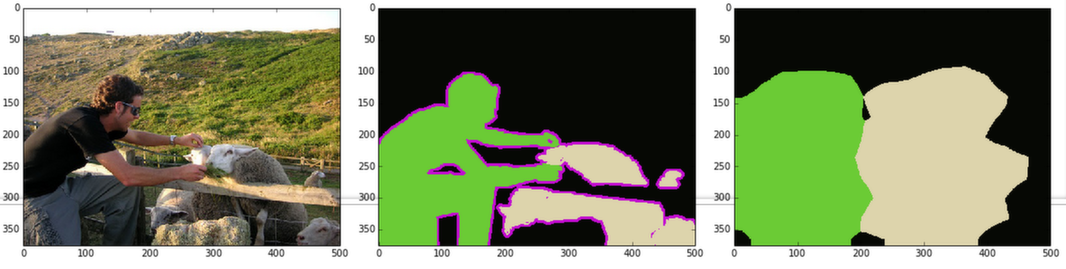}
\end{subfigure}\hfill
\begin{subfigure}[b]{.49\textwidth}
    \centering
    \includegraphics[width=\textwidth,height=2.5cm]{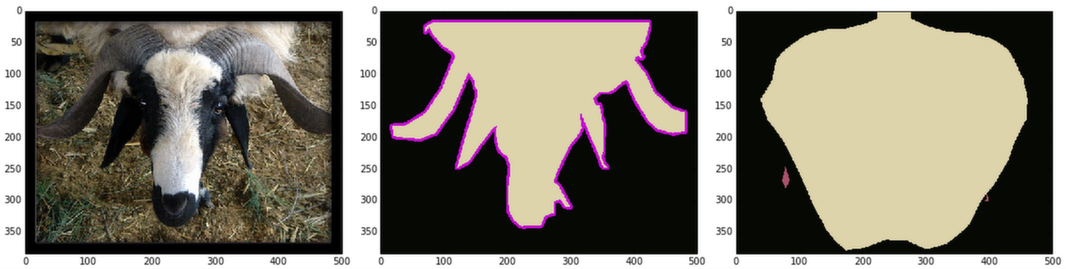}
\end{subfigure}\hfill

\begin{subfigure}[b]{.49\textwidth}
    \centering
    \includegraphics[width=\textwidth,height=2.5cm]{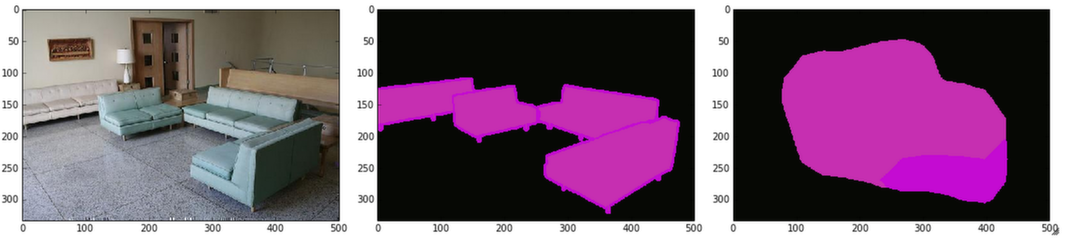}
\end{subfigure}\hfill
\begin{subfigure}[b]{.49\textwidth}
    \centering
    \includegraphics[width=\textwidth,height=2.5cm]{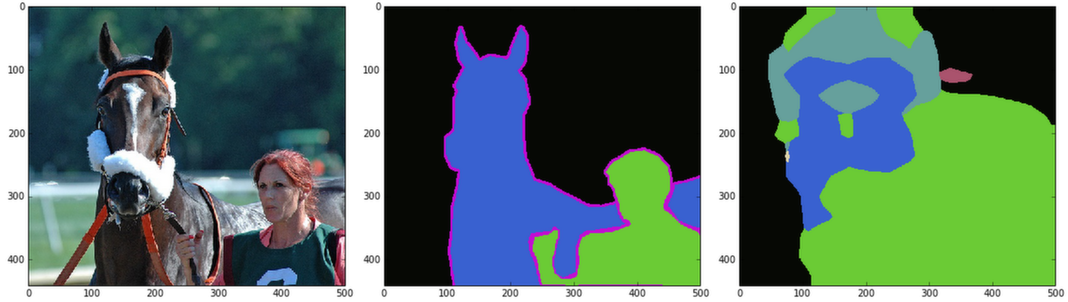}
\end{subfigure}\hfill

\end{center}
\caption{\emph{Sample images from PASCAL VOC 2011 val-segmentation data.} Each row shows input (left), ground truth (center) and MIL-FCN output (right).}
\label{fig:examples}
\end{figure}

Table~\ref{table:result} shows quantitative intersection-over-union (IU) scores while example outputs from MIL-FCN are shown in Figure~\ref{fig:examples}.
MIL-FCN achieves \textbf{96\% relative improvement} over the baseline results when the classifier is fine-tuned from the common classes.
These are preliminary but encouraging results.

% TODO is this right? early stopping + small step size?

% TODO I suggest titling this ``Discussion'' and
% (1) summarizing your workshop contribution
% (2) stating the one immediate next step to be taken
% ``future work'' can sometimes read as ``what I wish I had done by now''
% and that isn't so exciting.
%\section{Future Work}
\section{Discussion}

We propose a novel model of joint multiple instance and representation learning with a multi-class pixelwise loss inspired by binary MIL.
This model is learned end-to-end as a fully convolutional network for the task of weakly supervised semantic segmentation.
% TODO what does instance hypothesis mean? is that not the initial classifier output?
It precludes the need for any kind of proposal or instance hypothesis mechanisms.
Inference is fast ($\approx\textbf{1/5}$ \textbf{sec}).

These results are encouraging, and can be improved further.
Currently, the coarse output is merely interpolated; conditional random field regularization or super-pixel~\citep{achanta2012slic} projection could refine the predictions.
These grouping methods could likewise drive learning by selecting whole segments instead of single points for MIL training.
Moreover, controlling convnet learning by manipulating the loss map could have further uses such as encouraging consistency across images for co-segmentation or hard negative mining.
% TODO are you ok with releasing the model and code? if not, strike this out.
% The MIL-FCN model and code will be released open-source for investigating these possibilities.

%\subsubsection*{Acknowledgments}
%Trevor - Any acknowledgements ? Later.

\bibliography{iclr2015}
\bibliographystyle{iclr2015}

\end{document}